# Attention-Based LSTM Network for COVID-19 Clinical Trial Parsing

Xiong Liu[1], Luca A. Finelli[2], Greg L. Hersch[3], Iya Khalil[1]

[1]Data Science and AI, Novartis, Cambridge, MA, USA
[2]Data Science and AI, Novartis International AG, Basel, Switzerland
[3]Global Drug Development, Novartis Pharma AG, Basel, Switzerland
Email: xiong.liu@novartis.com

*Abstract*—COVID-19 clinical trial design is a critical task in developing therapeutics for the prevention and treatment of COVID-19. In this study, we apply a deep learning approach to extract eligibility criteria variables from COVID-19 trials to enable quantitative analysis of trial design and optimization. Specifically, we train attention-based bidirectional Long Short-Term Memory (Att-BiLSTM) models and use the optimal model to extract entities (i.e., variables) from the eligibility criteria of COVID-19 trials. We compare the performance of Att-BiLSTM with traditional ontology-based method. The result on a benchmark dataset shows that Att-BiLSTM outperforms the ontology model. Att-BiLSTM achieves a precision of 0.942, recall of 0.810, and F1 of 0.871, while the ontology model only achieves a precision of 0.715, recall of 0.659, and F1 of 0.686. Our analyses demonstrate that Att-BiLSTM is an effective approach for characterizing patient populations in COVID-19 clinical trials.

## I. Introduction

COVID-19 is an infectious disease caused by a new coronavirus [1]. As of September 26, 2020, more than 32.6 million coronavirus cases have been confirmed worldwide, with over 989,000 related deaths [2]. In the U.S. alone, there are more than 7 million coronavirus infections, with a death toll above 203,000 [2], [3]. Given the ongoing trend of the infection and a lack of effective interventions, pharmaceutical companies and research organizations are racing to develop life-saving treatments. According to ClinicalTrials.gov, there are 3438 studies for COVID-19 as of September 28, 2020 [4]. Only 10.7% of the studies (368) are completed. 76.6% of the studies (2634) are either still recruiting patients or not yet recruiting [4]. Best practices are not available for clinical trial design and patient recruitment remains a significant challenge.

Clinical trial planning with clear foresight is crucial. Fundamental requirements in clinical trial design include defining the research questions, endpoints, study arms, randomization, and eligibility criteria of the patient population [5]. Patient eligibility is often guided by the goal to generalize the results and minimize bias [5]. Overly restrictive patient selection has been reported to compromise study generalizability, reduce the benefit-cost ratio of clinical studies, and lead to the difficulty in interpreting and disseminating study results [6]. Recently, the FDA released new guidance on COVID-19 drug development, emphasizing the need for flexible trial designs and diverse patient populations [7].

Eligibility criteria play an essential role in defining the patient population or cohort. They describe the characteristics of the target populations of clinical studies. The criteria provide clinical practice guidelines for investigators to screen patients. The quality of criteria directly affects patient enrollment and study generalizability. The trial and error approach to define criteria often leads to many protocol amendments [6]. Feasibility assessment tools are needed to help investigators discover potential patient selection problems and make better eligibility criteria choices [8].

Clinical study design information is increasingly available publicly through public registries (e.g., ClinicalTrials.gov). In addition, the adoption of electronic health records, clinical data warehouses, and clinical data networks have generated enormous amounts of electronic patient data [6]. This has provided a new opportunity to transform the design of eligibility criteria using a data-driven approach.

Eligibility criteria are documented as unstructured free-text, which is not readily suitable for automated cohort definition and knowledge reuse or sharing [9], [10]. Natural language processing (NLP) methods can be used to parse free-text eligibility criteria into a structured and computable representation. The core NLP task is information extraction, which includes named entity recognition (NER) and entity attributes and relations extraction [10].

A number of NER methods have been developed for criteria parsing including rule-based methods and machine learning based methods. Rule-based methods use knowledge bases or ontologies to transform criteria text into computable concepts [11], [12]. Rule-based NER strongly relies on the terms and their interrelations defined in the ontology. It cannot identify new terms or entities. Machine learning methods can address the off-dictionary issue by learning from the way various concepts are used in written medical or life science language. For example, Kang et al. [13] annotated eligibility criteria from 230 Alzheimer's clinical trials to train conditional random field (CRF) models for entity recognition. However, traditional machine learning methods require sophisticated feature engineering to build accurate models.

More recently, deep learning methods have been introduced to enable more automatic and accurate extraction of entities in criteria parsing [14]. Deep learning is capable of understanding the semantic and syntactic relationship between words using a technique called "word embedding". Another competitive edge is from the deep neural architectures which can capture more context and semantics in the data. Deep learning can recognize terms and concepts not present in the ontology and reduce the overhead of feature engineering in traditional machine learning.

We explore a deep learning neural network, attention-based bidirectional Long Short-Term Memory (Att-BiLSTM), to extract entities from the eligibility criteria of COVID-19 trials. We leverage the annotation datasets described in [14] to train Att-BiLSTM models. We then apply an optimal model to all the COVID-19 eligibility criteria. To evaluate extraction quality, we develop an evaluation dataset from randomly sampled trials to measure the performance of Att-BiLSTM. We also compare its performance with that of ontology-based extraction using MeSH. We demonstrate that Att-BiLSTM outperforms the ontology-based method. This provides the rationale to choose the Att-BiLSTM extracted entities to represent the variables of cohort definition in COVID-19 trials.

## II. MATERIALS AND METHODS

### A. COVID-19 Dataset

We retrieved COVID-19 clinical trials registered in ClinicalTrials.gov as of August 15, 2020. The initial dataset included 3012 trials. We excluded trials with no eligibility text and trials without clear headings of "inclusion criteria:" and "exclusion criteria:", which led to 2998 trials. For each trial, we extracted the inclusion and exclusion criteria text and segmented it into sentences, where each sentence is a criterion. We obtained 27,352 criteria in total for the 2998 trials.

### B. Attention-based BiLSTM

Attention mechanism has gained popularity in image, speech and NLP fields [17], [18]. For NER task, attention mechanism has been introduced to enhance the BiLSTM model [14], [16].

We use a multi-layer BiLSTM based representation with attention over words [15]. Fig. 1 shows the architecture of attention-based BiLSTM for parsing a criterion. In the embedding layer, the criterion sentence is represented as a sequence of vectors $X = (x_1, x_2, \ldots, x_i, \ldots, x_n)$, where $x_i$ is the embedding vector of the word $i$, and $n$ is the length of the sentence. We use the word2vec vectors that were pre-trained on the trial descriptions and eligibility criteria of over 300K trials present in ClinicalTrials.gov [14].

Next, the embeddings are given as input to a BiLSTM layer. A forward LSTM computes a vector $\overrightarrow{h_i}$ to represent the word $i$ and its left context, and another backward LSTM computes a vector $\overleftarrow{h_i}$ to represent the word $i$ and its right context. Concatenating the two vectors yield one representation of the word $i$ in its context, $h_i$.

In the attention layer, an attention matrix A is used to calculate the similarity between the current target word and all words in the sentence. The attention weight value $a_{i,j}$ in the attention matrix is derived by comparing the $i$ th current target word representation $x_i$ with the $j$ th word representation $x_j$ in the sentence:

$$a_{i,j} = \frac{\exp(score(x_i, x_j))}{\sum_k \exp(score(x_i, x_k))}$$

Here, the score is defined as an alignment function with three alternatives including "Dot", "Multiply" and "Add". For Dot Attention, we simply dot the two representations together. For Multiply, we add in a single linear layer to Dot attention. For Add, we concatenate the representation together and then execute a 2-layer tanh multilayer perceptron (MLP).

Then a context vector $c_i$ is computed as weighted sum of each BiLSTM output $h_j$:

$$c_i = \sum_{j=1}^{n} a_{i,j} h_j$$

Next, the context vector and the BiLSTM output of the target word are concatenated as a vector to be fed to a tanh function to produce the output of attention layer $z_i$.

The CRF layer uses the $z_i$ as features to make tagging decisions for each output $y_i$. For a sequence of predictions $Y = (y_1, y_2, \ldots, y_n)$, its score is defined by the sum of transition scores and network scores:

$$s(X, Y) = \sum_{i=0}^{n} T_{y_i, y_{i+1}} + \sum_{i=1}^{n} P_{i, y_i}$$

$T$ is a tagging transition matrix such that $T_{i,j}$ represents the score of a transition from tag $i$ to tag $j$. $y_0$ and $y_n$ are the start and end tags of the sentence. $P$ is the matrix of scores output, where the element $P_{i,j}$ of the matrix is the score of the tag $j$ of the word $i$ in the sentence.

Then a softmax function is used to compute the conditional probability for the sequence $Y$ by normalizing the above score over all possible tag sequences $\tilde{Y}$:

$$p(Y|X) = \frac{e^{s(X,Y)}}{\sum_{\tilde{Y}} e^{s(X,\tilde{Y})}}$$

During training, the objective of the model is to maximize the log probability of the correct tag sequence. During inference, the best tag sequence (output) with the maximum score is given by:

$$Y^* = argmax\ s(X, \tilde{Y})$$

This is computed by dynamic programming [15].

### C. Ontology-based Entity Extraction

Ontology-based NER is a rule-based approach. It recognizes terms and concepts in unstructured text by checking against the pre-defined concepts and relationships. We use the MeSH ontology to scan text mentions of conditions, treatments, clinical measures and demographics. The model checks the standard name and synonyms of each ontological concept to find the matching terms in the criteria and automatically standardize the terms by assigning the concept label.

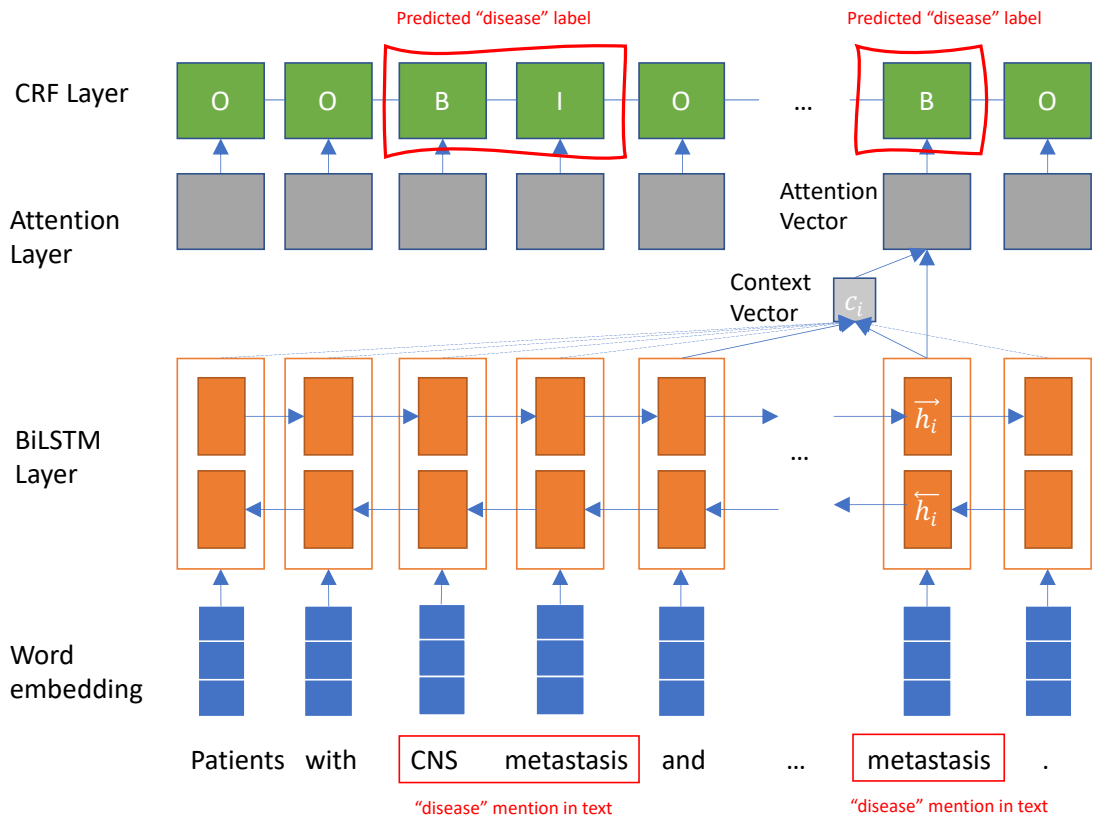

Fig. 1. The architecture of the attention-based BiLSTM model for criteria parsing. Figure adapted from [16].

### D. Entity Normalization

We normalize the extracted entities and use them as design variables to quantify clinical trials. The normalization process takes the following steps: 1) using fuzzy matching to link the extracted terms to MeSH ontologies; 2) using simple rules to standardize the non-matched terms, e.g., "pao2/fio2", "pao2/fio2 ratio", and "pao2 to fio2 ratio" are standardized to "pao2/fio2"; 3) using rule-based method to further standardize the same concept in heterogeneous formats, e.g., "hcq" and "hydroxychloroquine" are standardized to the same concept of "hydroxychloroquine".

## III. RESULTS

### A. Att-BiLSTM Models

We leverage the labeled dataset described in [14] to train NER models. This dataset includes 3,314 trials randomly sampled from ClinicalTrials.gov. We selected 11 entity types, including allergy, chronic disease, cancer, pregnancy, consent, treatment, clinical variables, language fluency, technology access, gender and age. This led to 92,937 entities for 42,542 criteria. This dataset becomes the training data. We randomly sampled 10% of the data as the test data used for performance measure.

As described in the Methods section, three attention methods (Dot, Multiply, Add) are designed for our Att-BiLSTM model. We trained the models with different attention methods on the training data using PyText [21]. The major hyper-parameters include LSTM dimension of 128, attention dimension of 64, MLP decoder hidden dimension of 256, dropout of 0.2, batch size of 64, and epochs of 10.

To select the best method, we tested the effect of these methods on the model performance. Table I shows the accuracy and loss of the attention-based models with different alignment functions on the test data.

TABLE I. PERFORMANCES OF DIFFERENT ATT-BILSTM MODELS ON THE TEST DATA

| Attention Method | Accuracy | Loss |
| --- | --- | --- |
| Multiply | 92.87 | 0.172 |
| Add | 92.62 | 0.174 |
| Dot | 92.52 | 0.181 |
| No attention | 92.55 | 0.179 |

The result shows that the model using the multiply attention method achieves the highest accuracy of 92.87% and lowest loss of 0.172. The dot attention achieves the lowest accuracy of 92.52% and highest loss of 0.181. The reason could be that the dot attention is just a transformation of the input without any trainable weights. We select the multiply attention model as our final model for prediction.

### B. Entity Extraction and Normalization

We use the selected Att-BiLSTM model to annotate the COVID-19 trials. We removed entities with a probability score less than 0.7 computed by Att-BiLSTM. Finally, 42,344 entities were extracted from 20,164 criteria for 2821 trials. We also use the MeSH ontology model to annotate the same set of trials for comparison.

Table II shows some samples of the extracted entities and the entity types. We can see that Att-BiLSTM deep learning result shows more details than the ontology modeling result. For example, given a term "known hypersensitivity to nivolumab", Att-BiLSTM recognizes it as an allergy, while the ontology model only recognizes part of the term "nivolumab" as a drug. Also, the ontology model does not recognize terms not defined in the MeSH ontology, such as immune checkpoint inhibitors, arterial oxygen saturation, and language fluency.

TABLE II. EXAMPLE ENTITIES AND TYPES

| Example | Att-BiLSTM annotation | Ontology-model annotation |
|---|---|---|
| Known hypersensitivity to nivolumab | Allergy | Drug |
| Patient with a history of thymoma | Chronic disease | Diseases |
| Active hematologic malignancy | Cancer | Diseases |
| Pregnant Women | Pregnancy | Persons |
| unwilling to practice a medically form contraception | Contraception Consent | Procedure |
| immune checkpoint inhibitors | Treatment | Not found |
| bone marrow transplantation | Treatment | Therapeutics |
| arterial oxygen saturation | Clinical variable | Not found |
| speaking and understanding English | Language fluency | Not found |
| people who are not internet accessible | Technology access | Not found |
| women | Gender | Persons |
| age over 18 years old | Age | Age groups |

### C. NER Performance Evaluation

We further evaluate the NER performance using a quantitative approach. We create a benchmark dataset by randomly sampling 10 COVID-19 trials and manually labeled 179 correct entities. We then use the dataset to evaluate the NER performance of Att-BiLSTM and the ontology model. Table III shows the performance of our models as evaluated on the 10 trial benchmark dataset.

BiLSTM achieves a precision of 0.942, recall of 0.810, F1 of 0.871, while MeSH ontology method only achieves a precision of 0.715, recall of 0.659, and F1 of 0.686.

TABLE III. PERFORMANCE MEASURE OF THE MODELS ON THE 10-TRIAL BENCHMARK DATASET

| Entity Recognition | Precision | Recall | F1 |
|---|---|---|---|
| Att-BiLSTM model | 0.942 (145/154) | 0.810 (145/179) | 0.871 |
| Ontology-based model | 0.715 (118/165) | 0.659 (118/179) | 0.686 |

Since the quality of Att-BiLSTM deep learning is higher, we use the Att-BiLSTM extracted entities rather than the ontology-based extracted entities. We then normalize the entity terms using the rule-based approached described in the Methods section. The normalized entities are used as variables to characterize clinical trials.

### D. Design Patterns

The extracted variables can be used to quantify clinical trials. We show one example to demonstrate the utility of the extracted variables to find design patterns across COVID-19 trials. We categorize the variable types by the medical conditions labeled by ClinicalTrials.gov. Fig. 2 shows the heat map on the frequency of variable types across conditions. The row is the variable type, the column is the COVID-19 related conditions, and the value is the unique count of trials. Note only variables that occur over 10 times are shown.

For COVID-19 specific studies, the most frequent entity types include chronic disease, treatment, age, pregnancy, and clinical variables. Other than COVID-19, acute respiratory distress syndrome and pneumonia are most frequent co-occurring conditions. For acute respiratory distress syndrome, the most frequent entity types include chronic disease, treatment, clinical variable, and pregnancy. For pneumonia, the most frequent entity types include chronic disease, treatment, pregnancy, and age. For all trials, there is a general tendency to consider chronic diseases and treatments.

We drill down the detailed variables for COVID-19 specific trials. The most frequent disease variables include HIV, pneumonia, respiratory failure, diabetes mellitus, and heart failure. The most frequent diagnosis and treatment variables include RT-PCR, mechanical ventilation, immunosuppressive agents, investigation therapies, and hydroxychloroquine. The most frequent clinical variables include oxygen saturation, spo2, respiratory rate, aspartate aminotransferases, and pao2/fio2.

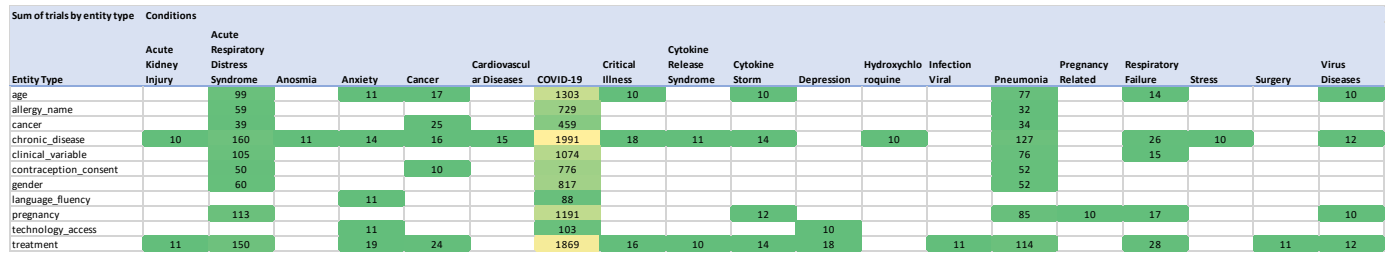

Fig. 2. Frequent eligibility criteria variables across COVID-19 and comorbidities

## IV. Discussions

Previous studies have tried to characterize COVID-19 trials in terms of treatments, single or double blinded design, randomization, and enrollment [19]. However, the authors are not aware of any studies that have focused on eligibility criteria analysis and cohort definition.

We demonstrate that entity extraction provides an effective way to quantify clinical trial parameters in eligibility criteria and that deep learning returns more detailed result than the ontology-based method.

The ontology-based method (e.g., using MeSH) suffers from finding terms/entities (variables) that are not defined in the dictionary, which leads to low recall. The deep learning Att-BiLSTM model can greatly improve the recall by finding off-dictionary terms/entities with finer granularity. For example, given a term "known hypersensitivity to nivolumab", Att-BiLSTM recognizes it as an allergy, while the ontology model only recognizes the term "nivolumab" as a drug.

The extracted entities/variables can be used for knowledge reuse and downstream machine learning applications such as predicting trial success rates [22]. We demonstrated a utility of the variables to find common design variables or patterns across trials. The most common disease variables include HIV, pneumonia, and respiratory failure; the most frequent diagnosis and treatment variables include RT-PCR, mechanical ventilation, and immunosuppressive agents; and the most frequent clinical variables include oxygen saturation, spo2, respiratory rate, aspartate aminotransferases, and pao2/fio2.

These results correspond with previous findings. Recent studies have shown that pneumonia and acute respiratory distress syndrome are common comorbidities of COVID-19 [23], [24]. RT-PCR is a common diagnosis method to confirm COVID-19 [25], [26]. A recent work on the COVID-19 severity index (CSI) shows that oxygen requirement is an important measure of critical illness [20].

The results also reveal the frequent design variables in co-occurring conditions or comorbidities of COVID-19, such as acute respiratory distress syndrome and pneumonia. For all trials, there is a tendency to consider chronic diseases and treatments. This may imply that COVID-19 related clinical studies are very careful about selecting patient populations by considering patient history of chronic diseases and prior treatments.

Our study has limitations in modeling the attributes of design variables, such as numeric values, negation, and temporal constraints. To support more detailed eligibility assessment for clinical studies, it is necessary to model all possible eligibility variables and all possible values. Therefore, more sophisticated information extraction models will be required. These variables and values will provide a more structured representation of the criteria, which unlocks the potential for more advanced pattern analysis and machine learning tasks.

Another limitation is that we only used MeSH as an example ontology for the ontology-based model. Other ontologies such as UMLS and NCI Thesaurus will need to be tested.

## V. Conclusions

In this study, we investigated the performances of different versions of Att-BiLSTM models to extract design variables from the COVID-19 eligibility criteria. The multiply attention method achieves the best performance with an accuracy of 92.87, while the dot attention method achieves the worst performance with an accuracy of 92.52. BiLSTM without attention achieves an accuracy of 92.55. This shows the impact of attention methods on the BiLSTM model performance.

Our evaluation on a benchmark data shows that Att-BiLSTM outperforms the ontology model in parsing COVID-19 trials. Att-BiLSTM achieves a F1 of 0.871, while the MeSH ontology method only achieves a F1 of 0.686. Qualitative analysis also shows that Att-BiLSTM can extract more detailed entities than the ontology model.

The extracted eligibility criteria variables provide a mechanism for characterizing patient populations eligible in COVID-19 trials. We used the extracted variables to find design patterns in COVID-19 eligibility criteria. Our result not only confirmed previous findings about COVID-19, but also discovered new knowledge about comorbidities. In summary, our study demonstrated the effectiveness of Att-BiLSTM deep learning in protocol criteria parsing.